\documentclass[11pt]{article}

\usepackage{geometry}
\usepackage[utf8]{inputenc} 
\usepackage[T1]{fontenc}    
\usepackage{hyperref}       
\usepackage{url}            
\usepackage{booktabs}       
\usepackage{amsfonts}       
\usepackage{nicefrac}       
\usepackage{microtype}      
\usepackage{xcolor}         

\usepackage{graphicx}
\usepackage{mathtools}
\usepackage{dsfont}
\usepackage{amsmath, amsthm, amssymb}

\title{Minima and Critical Points of the Bethe Free Energy Are Invariant Under Deformation Retractions of Factor Graphs}

\usepackage{graphicx}
\usepackage{mathtools}
\usepackage{dsfont}
\usepackage{amsmath, amsthm, amssymb}
\usepackage{authblk}
\usepackage{cite}

\author[1,2]{Grégoire Sergeant-Perthuis\thanks{\texttt{gregoire.sergeant-perthuis@sorbonne-universite.fr}}}
\author[1]{Léo Boitel}

\affil[1]{CQSB, Sorbonne Université, Paris, France}
\affil[2]{Ouragan, Inria Paris, France}

 \theoremstyle{plain}
 \newtheorem{defn}{Definition}
  
  \newtheorem*{defn*}{Definition}
 
  \theoremstyle{plain}
  \newtheorem{thm}{Theorem}
  
  \newtheorem{prop}{Proposition}
  \newtheorem*{prop*}{Proposition}
  \newtheorem{lem}{Lemma}
  \newtheorem{cor}{Corollary}
   \newtheorem*{cor*}{Corollary}
  \newtheorem*{theo*}{Theorem}
  \newtheorem*{thm*}{Theorem}
  
  \theoremstyle{remark}
  \newtheorem{rem}{Remark}
 
 \usepackage[toc,page]{appendix}
\usepackage{calrsfs}
\usepackage{tikz-cd}

\newcommand{\R}{\mathbb{R}}
\newcommand{\A}{\mathcal{A}}
\newcommand{\B}{\mathcal{B}}

\usepackage{pstricks,pst-node,pst-tree}

\usepackage{bm}

\usepackage{relsize}

 \usepackage{xcolor}

\newcommand{\im}{\operatorname{im}}

\theoremstyle{plain}

\numberwithin{equation}{section}

\newcommand{\id}{\text{id}}

\usepackage{hyperref} 

\date{}
\begin{document}

\maketitle

\begin{abstract}
In graphical models, factor graphs, and more generally energy-based models, the interactions between variables are encoded by a graph, a hypergraph, or, in the most general case, a partially ordered set (poset). Inference on such probabilistic models cannot be performed exactly due to cycles in the underlying structures of interaction. Instead, one resorts to approximate variational inference by optimizing the Bethe free energy. Critical points of the Bethe free energy correspond to fixed points of the associated Belief Propagation algorithm. A full characterization of these critical points for general graphs, hypergraphs, and posets with a finite number of variables is still an open problem. We show that, for hypergraphs and posets with chains of length at most 1, changing the poset of interactions of the probabilistic model to one with the same homotopy type induces a bijection between the critical points of the associated free energy. This result extends and unifies classical results that assume specific forms of collapsibility to prove uniqueness of the critical points of the Bethe free energy.

\end{abstract}

\section{Introduction}

Graphical models are probabilistic models that account for conditional independence relations between variables, derived from the connectivity properties of substructures of a graph~\cite{Lauritzen}.

 The Hammersley-Clifford theorem relates those conditional independence properties to a a way the joint distribution should factor: it is the product of factor, on per clique of the graphs, each factor depending only on variables inside a given clique. Factor graphs are a generalization of graphical models which encode how distribution factor into terms that depend on sub collection of variables. In applications, some of the variables in a factor graph are called \emph{hidden} and others \emph{observed}; computing the posterior of the joint distribution that factorizes according to the graph, conditioned on the observed variables, is computationally intractable when done naively. One instead resorts to an \emph{approximate inference} method, which corresponds to minimizing the Bethe Free Energy under linear constraints. The Belief Propagation algorithm is one algorithm used to find the critical points of the free energy: critical points of the approximate free energy are in correspondence with the fixed points of the algorithm. Therefore, one way to study the critical points of the Bethe free energy is to study the fixed points of the respective Belief Propagation algorithm. These critical points have been observed to be non-unique, and characterizing the set of such points is still ongoing ~\cite{4385778,10.5555/2073876.2073935,heskes2002stable,ihler2005loopy,mooij2012sufficient,dahlen2012message,martin2012local,knoll2017fixed,9852264,furtlehnerloopy,Zivan_Lev_Galiki_2020,grim2022message,peltreconf}.

The topology of the graph or hypergraph encoding the dependencies of the variables affects the possible set of critical points one can obtain. When the graph underlying the graphical model is acyclic, i.e., possesses no cycles, it is well known that the Bethe free energy has only one critical point, and that approximate inference in this case is exact. In Section 4.2.1 of \cite{wainwright2008graphical}, hypertrees or acyclic hypergraphs are introduced. In \cite{welling2012choice}, the authors study how pruning the poset can keep the free energy unchanged, and in Section 6.1.1 of \cite{Peltre}, a notion of retractable hypergraph is given for which the free energy has only one critical point. Although the operations studied and conditions imposed are of a topological nature, they seem to be dispersed results focused on very particular aspects of the poset. Understanding the impact of the topology of finite factor graphs with cycles on the critical points of the Bethe free energy remains to be explored in a more systematic manner. We aim to address this by characterizing the critical points of this energy in terms of the homotopy type of the hypergraph associated with the factor graph.
In this article, we show that, for a good notion of homotopy on factor graphs, the critical points of the Bethe Free Energy are isomorphic.

\section{The Bethe Free Energy and the General Belief Propagation algorithms}

\subsection{Hypergraphs, Factor Graphs, and Posets}

Let $I$ be a finite index set of random variables: for each $i \in I$, there is a variable $X_i$ taking values in a finite set $E_i$. A hypergraph denoted as $\mathcal{H}=(I,A)$ is a finite collection of vertices $I$, and a finite collection $A$ of subsets $a\subseteq I$ called hyperedges, i.e. $A$ is a subset of the powerset $\mathcal{P}(I)$.

Let us denote by \(\R_{\ge 0}\) (respectively \(\R_{>0}\)) the set of non-negative (respectively strictly positive) real numbers.

\begin{defn}[Factor graph \cite{kschischang2002factor,yedidia2005}]
A factor graph $\mathcal{F}=(\mathcal{H},E,f)$ is made of a hypergraph $\mathcal{H}=(I,A)$, a collection of finite sets $(E_i; i\in I)$ and a collection of factor $f_a:\prod_{i\in a}E_i\to \R_{\geq 0}$, one for each $a\in A$. In this document, we will call factor graph a couple $\mathcal{F}=(\mathcal{H},E,H)$, where for each $a\in \A$, $H_a:\prod_{i\in a}E_i\to \R$ is a function that is not necessarily positive; to each $H_a$ we associate the factor $f_a=e^{-H_a}$ which is a strictly positive function. Our convention for factor graphs is more restrictive as we assume that each factor is a strictly positive function, however it is a common assumption \cite{yedidia2005}. 
\end{defn}

The standard way to present factor graphs is to introduce a bipartide graph made of variable nodes, $i\in I$, and of factor nodes, $a\in A$, where each factor node is labelled with $f_a$. For all $i\in I$, and $a\in A$, if $i\in a$ then one adds a directed edge $a\to i$. 
A hypergraph is a particular example of poset $\A(\mathcal{H})$, which elements are $I\sqcup A$ and relation $i\leq a$ are generated by $i\in a$ with $i\in I$ and $a\in A$. And the Hasse diagram of~\(\A(\mathcal{H})\) is the directed bipartite graph associated to the factor graph with all arrows reversed, i.e.\ where every arrow \(a \to i\) is replaced by \(i \to a\).

\subsection{Bethe Free Energy}\label{Bethe-free-general}
Let $I$ be a finite set, let $\mathcal{A} \subseteq \mathcal{P}(I)$ be a collection of subsets of $I$. For each $a \in \mathcal{A}$, consider a factor function $f_a: E_a \to \mathbb{R}_{\geq 0}$, where $E_a = \prod_{i \in a} E_i$ and $E_I=\prod_{i\in I} E_i$. We denote by \(\mathbb{P}(E_a)\) the space of probability distributions over \(E_a\); that is, \(P \in \mathbb{P}(E_a)\) is a collection $(P(x); x\in E_a)$ such that $\sum_{x\in E_a} P(x) = 1$. Similarly let $P\in \mathbb{P}_{>0}(E_a)$ when $P\in \mathbb{P}(E_a)$ and for all $x\in E_a$, $P(x)>0$.

A joint distribution $P \in \mathbb{P}(E_I)$ is said to factor according to the collection $\{f_a\}_{a \in \mathcal{A}}$ if, for any $x \in E_I$,
\[
P(x) = \prod_{a \in \mathcal{A}} f_a(x_a),
\]
where $x_a = (x_i)_{i \in a}$ denotes the restriction of $x$ to the coordinates indexed by $a$.
We denote the projection from \(x \in E_a\) to \(x_b \in E_b\), with \(b \subseteq a\), as \(\pi^a_b\).

The M\"obius function of a poset $\mathcal{A}$ (Proposition 2 in \cite{Rota}) is the unique collection $\mu = \big(\mu(a, b) \in \mathbb{Z} ; b, a \in \mathcal{A} \text{ such that } b \leq a\big)$ satisfying, for any $a, c \in \mathcal{A}$ with $c \leq a$,
\[
\delta_{a,c} = \sum_{b: c \leq b \leq a} \mu(b, c) = \sum_{b: c \leq b \leq a} \mu(a, b).
\]

\(\delta_{a,c} = 1\) if \(a = c\), and \(0\) otherwise. We will also use the notation \(1[C(x)]\) for the indicator function which equals \(1\) if the condition \(C(x)\) is satisfied and \(0\) otherwise.

The \emph{Generalized Bethe free energy} \cite{yedidia2005, PeltrePhD} associated with the pair $(\mathcal{A}, (f_a)_{a \in \mathcal{A}})$ is defined for any collection of probability distributions $Q = (Q_a \in \mathbb{P}(E_a))_{a \in \mathcal{A}}$ as
\begin{equation}\label{bethe-free-energy}
BT_{\A,H}(Q) = \sum_{a \in \mathcal{A}} c(a) \left( \mathbb{E}_{Q_a}[H_a] - S(Q_a) \right),
\end{equation}
where for each $a \in \mathcal{A}$,
\[
c(a) = \sum_{b \geq a} \mu(b, a),
\qquad
H_a(x_a) = \sum_{b \subseteq a} -\ln f_b(x_b).
\]

We call \(H_a\) a Hamiltonian, it is simply a function \(H_a : E_a \to \R\). We recover the classical Bethe approximation for graphical models when $\mathcal{A} = \mathcal{A}(G)$ for an undirected graph $G$ (seen as a hypergraph), and for factor graphs when $\mathcal{A} = \mathcal{A}(\mathcal{H})$ for a hypergraph $\mathcal{H}$ (see Chapter 4 \cite{wainwright2008graphical}). The relevant family of distributions 
\(\left(Q_a \in \mathbb{P}(E_a))_{a \in \A}\right)\) 
is the one satisfying marginalization consistency: for all \(b \subseteq a\), we require that \(Q_b\) is the marginal of \(Q_a\), obtained by summing over the configurations of the variables $X_{a \setminus b} = (X_i \;|\; i \in a,\; i \not\in b)$. This compatibility condition is called in \cite{wainwright2008graphical} the \emph{polyhedral outer bound} of the marginal polytope, and is denoted by \(\mathbb{L}(\A)\). More precisely,

\begin{equation}
\mathbb{L}(\A)
:= \left\{\, Q = (Q_a)_{a \in \mathcal{A}} \in \prod_{a\in \mathcal{A}} \mathbb{P}(E_a)\,\middle|\,
\forall\, b \le a,\ \forall\, x \in E_b,\;
Q_b(x) = \sum_{\substack{y \in E_a\\ y_{b} = x}} Q_a(y)
\right\}.
\end{equation}

The marginal polytope corresponds to all the probability distributions $(Q_a,a\in \A)$ that are the marginals of a global $P\in \mathbb{P}(E_I)$. $\mathbb{L}(\A)$ is called the polyhedral outer bound, as it is a polytope defined by linear equalities and inequalities that contains strictly the marginal polytope: there may exist a collection of local distributions $(Q_a)_{a \in \mathcal{A}}$ that do not arise from any global distribution $Q \in \mathbb{P}(E)$. 
Let us first remark that $\mathbb{L}(\mathcal{A})$ is a compact set, and the Generalized Bethe free energy is continuous on this set. Therefore, it has at least one global minimum.

The interior of \(\mathbb{L}(\A)\), denoted \(\mathbb{L}(\A)^\circ\), consists of distributions \(Q_a \in \mathbb{P}_{>0}(E_a)\), for \(a \in \A\), that are compatible by marginalization. Characterizing the critical points of the Generalized Bethe free energy within \(\mathbb{L}(\A)^\circ\) is an important problem in the study of graphical models. In the next section, we will recall and reprove a known result: there is always at least one critical point in this set when the poset \(\A\) arises from a factor. The authors found it difficult to present a full and simple proof of this result; that is why we chose to prove the result in the article. We do so by characterizing these points as fixed points of the Generalized Belief Propagation (GBP) algorithm. The proof relies on Brouwer's fixed-point theorem applied to Generalized Belief Propagation (GBP), analogous to its application in the max-product version of Belief Propagation \cite{wainwright2004tree}.
However, for general posets (of rank, with maximal chains that can be of length greater than 2), this result does not apply. Although GBP maps the interior of \(\mathbb{L}(\mathcal{A})\) into itself, it cannot be continuously extended to the boundary, making BP not a continuous operator from \(\mathbb{L}(\mathcal{A})\) to itself.

Hamiltonians \((H_a)_{a\in \A}\) and factors \((f_a)_{a\in \A}\) are related by the zeta function of the poset \(\A\): \(H_a = \sum_{b \subseteq a} -\ln f_b\); this operator has an inverse given by \(-\ln f_a = \sum_{b \subseteq a} \mu(b, a) H_b\). Starting from \(\ln f\) or \(H\) is equivalent; in what follows, we consider Hamiltonians instead of factors.

\subsection{The General Belief Propagation algorithm, and existence of critical points of the Bethe Free energy for factor graphs }

\subsection{The general setting}
The General Belief Propagation algorithm is used to find the critical points of the Bethe Free Energy as it is a well know fact that the critical point of the Bethe Free Energy are in correspondance with fix points of the General Belief Propagation algorithm \cite{yedidia2005,PeltrePhD}; for a detail proof of this result see Proposition 3 \cite{sergeantperthuis2025functorialitybeliefpropagationalgorithms}. 

Let $I$ be a finite set that serves as index for variables $(X_i\in E_i; i\in I)$, each of which takes values in a finite set $E_i$; let $\A \subseteq \mathcal{P}(I)$ be a collection of subsets of I. Let $(H_a:E_a\to \mathbb{R}, a \in \A)$ be a collection of Hamiltonians. Let us denote the update rule of the General Belief Propagation algorithm as $\text{BP}$. $\text{BP}$ acts on messages that we will now define. In the classical presentation of the algorithm, there are two types of messages at each time $t \in \mathbb{N}^{\ast}$. For elements $a, b \in \mathcal{A}$ such that $b \subseteq a$, we have top-down messages $m_{a \to b} \in \mathbb{R}_{>0}^{E_b}$ and bottom-up messages $n_{b \to a} \in \mathbb{R}_{>0}^{E_b}$. 

These messages are related as follows:

\begin{equation}\label{GBP1}
\forall a,b\in \A, \text{s.t. }  b\subseteq a, \quad n_{b \to a}^t= \prod_{\substack{c: b\subseteq c\\ c\not \subseteq  a}}m_{c\to b}^t
\end{equation}

Beliefs, which are interpreted as probability distributions up to a multiplicative constant are defined as follows:

\begin{equation}\label{GBP2}
\forall a \in \A,\forall x_a\in E_a \quad b_a^t(x_a)\propto e^{-H_a(x_a)}\prod_{\substack{b\in \A:\\ b\subseteq a}} n_{b\to a}^t(x_b)
\end{equation}

where $\propto$ stands for proportional to.
The multiplication of function $n_{b\to a}$ that have different domains is made possible because there is an the embedding of $\R^{E_b}$ into $\R^{E_a}$ implicitly implied in the last equation; indeed, for $x \in E_a$ and $f \in \R^{E_b}$, $f: x \mapsto f(x_b)$ defines a function from $E_a$ to $\mathbb{R}$.

For simplicity, we require that $b_a$ be a probability distribution and normalize it accordingly. The update rule is given by,

\begin{equation}\label{GBP3}
\forall a,b\in \A \text{s.t.} b\subseteq a, \forall x\in E_b, \quad m_{a\to b}^{t+1}(x)= m_{a\to b}^{t}(x)\frac{\sum_{y\in E_a: y_b=x} b_a^t(y)}{b_b^t(x)}
\end{equation}

One observes that in the previous Equation~\ref{GBP3}, any normalization of beliefs does not change the update rule.

 The update rule can be rewritten in a more condensed manner, updating only the top-down messages, for all $a,b\in \A$, such that $b\leq a$,

\begin{align}\label{BP-update-rule}
 m_{a\to b}^{t+1}(x)&=  m_{a\to b}^t(x)\frac{\sum_{\substack{y\in E_a: y_b=x}} e^{-H_a(y)} \prod_{\substack{c\in \A:\\ c\subseteq a}}\prod_{\substack{d: c\subseteq  d\\ d\not \subseteq a}}m_{d\to c}^t (y_c)}{e^{-H_b(x)}\prod_{\substack{c\in \A:\\ c\subseteq b}}\prod_{\substack{d: c\subseteq  d\\ d\not \subseteq b}}m_{d\to c}^t (x_c)}
\end{align}

We denote the collection $(m_{a \to b}; a,b\in \A, b\leq a)$ as $m$. We denote $\text{BP}: \prod_{a,b: b \subseteq a} \mathbb{R}^{E_b} \to \prod_{a,b: b \subseteq a} \mathbb{R}^{E_b}$ as the operator underlying the update rule of Equation \ref{BP-update-rule}, i.e., we define $\text{BP}(m^t) = m^{t+1}$.

Consider the collection $\left(C_{a\to b} m_{a\to b}; a,b\in \A, b\leq a\right)$, where $C_{a\to b}$ is a strictly positive constant, i.e., it does not depend on $x\in E_b$. Then, there is a collection of constants $(C'_{a\to b} > 0; a,b\in \A, b\leq a)$ such that  
\[
\text{BP}(C_{a\to b} m_{a\to b};b\leq a) = (C'_{a\to b}\cdot \text{BP}(m)_{a\to b}; b\leq a).
\]

Furthermore, the associated beliefs defined by Equation~\ref{GBP2} remain unchanged under multiplication of $m_{a\to b}$ by a constant $C_{a\to b}$ for all $a,b\in \A$ such that $b\leq a$.

Therefore, $\text{BP}$ is an algorithm that preserves the equivalence classes $\{C \cdot m\}$, i.e., it is defined by the relation $m \sim m'$ whenever there is a collection of scalars $(C_{a\to b} \neq 0; a,b\in \A, b\leq a)$ such that  
\[
m_{a\to b} = C_{a\to b} m'_{a\to b} \quad \text{for any } a,b\in \A \text{ with } b\leq a.
\]
We shall denote the equivalence class of $m$ as $[m]$. The action of $\text{BP}$ on the equivalence classes of messages is denoted by $[\text{BP}]$ and defined as $[\text{BP}]([m]) = [\text{BP}(m)]$.

\subsection{The special case of factor graphs}

The update rule in Equation \ref{BP-update-rule} simplifies in the context of factor graphs, as we will explain now. Let $\mathcal{F}$ be a factor graph and $\mathcal{H}=(I,A)$ the associated hypergraph. We denote by $\mathcal{N}(a)$ the set of vertices $i \in I$ such that $i \in a$, and $\mathcal{N}(i)$ the set of hyperedges $a$ such that $i \in a$. Note that any $m^{t}_{a \to a}$ are kept unchanged over time and do not appear in the update rule; one can choose them to be equal to $1$, or simply omit them. The only message we must consider are those on arrows $ a\in A\to i\in I$;  remark that the part of the numerator of the update rule can be written as:

\begin{align}
\prod_{\substack{c \in \mathcal{A} \\ c \subseteq a}}
\;\prod_{\substack{d : c \subseteq d \\ d \not\subseteq a}}
m_{d \to c}^t(x_c)
&=
\prod_{j \in \mathcal{N}(a)} \;\prod_{b \in \mathcal{N}(j) \setminus a} m_{b \to j}^t(x_j)\\
&=\prod_{b \in \mathcal{N}(i) \setminus a} m_{b \to i}^t(x_i)\cdot\prod_{j \in \mathcal{N}(a)\setminus i} \;\prod_{b \in \mathcal{N}(j) \setminus a} m_{b \to j}^t(x_j)  
\end{align}

 \(\mathcal{N}(j)\setminus a\) denotes the set \(\mathcal{N}(j)\) without the element \(a\); this is an abuse of notation for \(\mathcal{N}(j)\setminus\{a\}\).  Similarly for \(\mathcal{N}(a)\setminus j\).

Part of the denominator can be rewritten as:

\begin{equation}
\prod_{\substack{c \in \mathcal{A} \\ c \subseteq i}}
\;\prod_{\substack{d : c \subseteq d \\ d \not\subseteq i}}
m_{d \to c}^t(x_c)=\prod_{b\in \mathcal{N}(i)}
m_{b \to i}^t(x_i)
\end{equation}

Remark that the numerator simplifies as,

\begin{equation}
    m_{a\to i}^t(x_i)\prod_{b \in \mathcal{N}(i) \setminus a} m_{b \to i}^t(x_i)= \prod_{b\in \mathcal{N}(i)}
m_{b \to i}^t(x_i)
\end{equation}
and this previous term only depends on $x_i$ and therefore factors through the sum in the numerator. In the end the update rule of $\text{BP}$ for factor graphs is simply,

\begin{align}\label{BP-update-rule-factor-graph}
 m_{a\to i}^{t+1}(x_i)&=  e^{H_i(x_i)}\sum_{\substack{y\in E_a: y_i=x_i}} e^{-H_a(y)} \prod_{j \in \mathcal{N}(a)\setminus i} \;\prod_{b \in \mathcal{N}(j) \setminus a} m_{b \to j}^t(y_j)
\end{align}

Let us now remark that if all the messages $m_{a\to i}^t > 0$ for $a \in A$ and $i \in a$, then the messages $m_{a\to i}^{t+1} > 0$ are also strictly positive at time $t+1$. In particular, the associated beliefs are given by
\begin{equation}
   \forall a \in A, \ \forall x_a \in E_a, \quad  
   b_a^t(x_a) \propto e^{-H_a(x_a)} 
   \prod_{i \in a} \prod_{\substack{b: b \neq a \\ i \in b}} m_{b\to i}^{t}(x_i).
\end{equation}

and 
\begin{equation}
   \forall a \in I, \ \forall x_i \in E_i, \quad  
   b_i^t(x_i) \propto e^{-H_i(x_i)} 
\end{equation}

are strictly positive, i.e., for all $a \in \A$ and $x_a \in E_a$, we have $b_a(x_a) > 0$. For strictly positive messages, the update rule in Equation~\ref{BP-update-rule-factor-graph} can be rewritten using $\ln m_{a \to i}^t$ denoted as $M_{a\to i}$, the associated update rule is, 

\begin{equation}\label{BP-log-update-rule-factor-graph}
M_{a \to i}^{t+1}(x_i) 
= H_i(x_i) 
+ \ln \sum_{\substack{y\in E_a :y_i = x_i}} 
    \exp\left(
        -H_a(y_a)
        + \sum_{j \in \mathcal{N}(a) \setminus i} 
            \sum_{b \in \mathcal{N}(j) \setminus a} 
                M_{b \to j}^t(y_j)
    \right).
\end{equation}

We denote the update rule of the previous equation, Equation~\eqref{BP-log-update-rule-factor-graph}, as $\log \text{BP}$, i.e.,
\[
   M_{a \to i}^{t+1}(x_i) = \bigl(\log \text{BP}(M)\bigr)_{a \to i}(x_i).
\]

As discussed previously, $\text{BP}$ is defined on messages up to a multiplicative constant for each message. Similarly, $\log \text{BP}$ is defined up to an additive constant. We denote the associated equivalence class by $\left[ \,\cdot\, \right]$.
Let us recall the notation: for any $a \in A$ and $i \in I$ such that $i \in a$,  
\[
C_{a \to i}(M) = \sup_{x_i \in E_i} \log \text{BP}_{a \to i}(M).
\]  
In particular,  
\[
\log \text{BP}(M - C(M)) = \log \text{BP}(M) + C(M),
\qquad
[\log \text{BP}(M)] = [\log \text{BP}(M) - C(M)].
\]  
Note that, by definition of $C$, for any $a \in A$ and $i \in I$ such that $i \in a$,  
\[
\sup_{x_i \in E_i} \Big( \log \text{BP}(M)_{a \to i} - C(M)_{a \to i} \Big) = 0.
\]
\begin{prop}\label{fix-point-BP}
The Belief propagation algorithm up to multiplicative constants has at least one fixed point such that all messages are strictly positive, i.e. there is a collection $\left(M^*_{a\to i}\in \mathbb{R}^{E_i}; a\in A,i\in I, \text{ s.t. } i\in a \right)$ such that,

\begin{equation}
    \left[M^*\right]=\log \text{BP}(\left[M^*\right])
\end{equation}

\end{prop}

\begin{proof}
The proof of the existence of fixed points of the max-product algorithm on graphs \cite{wainwright2004tree} can be adapted into a proof of the existence of fixed points of the Belief Propagation algorithm on factor graphs. \\

One remarks that, 

\[
x \in E_a \;\;\mapsto\;\; 
\sum_{j \in \mathcal{N}(a) \setminus i} 
\;\sum_{b \in \mathcal{N}(j) \setminus a} 
m_{b \to j}^t(x_j)
\]
does not depend on $x_i$. Let us denote this terms as $g(x_{a\setminus i})$. Therefore, for any two $x, z \in E_i$, 
\begin{align}
    \log \text{BP}(M)_{a \to i}(x)
= H_i(z) +& \bigl(H_i(x) - H_i(z)\bigr)+
\\
&\ln \sum_{\substack{y \in E_a \\ y_i = x_i}} 
    \exp\left(
        -H_a(y_{a \setminus i}, z) 
        + \bigl(H_a(y_{a \setminus i}, x) - H_a(y_{a \setminus i}, z)\bigr)
        + g(x_{a \setminus i})
    \right).
\end{align}

Therefore, 

\begin{equation}
\log \text{BP}(M)_{a \to i}(x)
\leq \log \text{BP}(M)_{a \to i}(z)+ \sup_{x,z\in E_i}\vert H_i(x)-H_i(z)\vert +\sup_{y,y_1\in E_a} \vert H_a(y)-H_a(y_1)\vert
\end{equation}

By exchanging $x$ and $z$, one obtains that

\begin{equation}
\sup_{x,z\in E_i}\vert \log \text{BP}(M)_{a \to i}(x)- \log \text{BP}(M)_{a \to i}(z)\vert \leq \sup_{x,z\in E_i}\vert H_i(x)-H_i(z)\vert +\sup_{y,y_1\in E_a} \vert H_a(y)-H_a(y_1)\vert
\end{equation}

Therefore there is $\Lambda$ such that

\begin{equation}\label{lemme-bound}
\sup_{\substack{a \in A,\, i \in I \\ i \in a}} 
\;\sup_{x,y \in E_i} 
\big| \log \text{BP}(M)_{a \to i}(x) - \log \text{BP}(M)_{a \to i}(y) \big| 
\;\leq\; \Lambda.
\end{equation}

In particular, one can deduce from Equation \ref{lemme-bound} and the fact that  

\[
\sup_{x_i \in E_i} \Big( \log \text{BP}(M)_{a \to i}(x) - C(M)_{a \to i} \Big) = 0,
\]

that,  

\begin{equation}
\sup_{\substack{a \in A \\ i \in a}} \ \sup_{x \in E_i} 
\big| \log \text{BP}(M)_{a \to i}(x) - C(M)_{a \to i} \big| \leq \Lambda.
\end{equation}

Denote by $K$ the subset of messages that are bounded by $\Lambda$. In other words,  
\begin{equation}
K := \left\{ M \ \Bigg| \ 
\sup_{\substack{a \in A \\ i \in a}} \ \sup_{x \in E_i} 
\big| M_{a \to i}(x) \big| \leq \Lambda \right\}.
\end{equation}
And denote by $[K]$ the set of equivalence classes,  
\begin{equation}
[K] := \{ [M] \mid M \in K \}.
\end{equation}

Then, $[\log \text{BP}]$ maps $[K]$ into $[K]$.  
Moreover, $[K]$ is a convex and compact set.  
Therefore, by Brouwer's fixed-point theorem, $[\log \text{BP}]$ admits at least one fixed point, denoted by $[M^*]$.

\end{proof}

\begin{cor}
Let $E = \prod_{i \in I} E_i$ and let $\mathcal{H}$ be a factor graph.  
Let $\mathcal{A}$ be the associated poset of $\mathcal{H}$.  
Then the generalized Bethe free energy $BT_{\mathcal{A},\mathcal{H}}$ admits at least one critical point in the interior of the polyhedral outer bound of the marginal polytope $\mathbb{L}(F)$.
\end{cor}

\begin{proof}
Let $\mathcal{H} = (I, A)$ be a hypergraph with nodes $I$ and hyperedges $A$.  
By Proposition~\ref{fix-point-BP}, there exists a fixed point $[M^*]$ of $[\log \text{BP}]$.  
For any $a \in A$ and $i \in I$ such that $i \in a$, denote
\[
m^*_{a \to i} \propto e^{M^*_{a \to i}}.
\]
Let $b$ be the associated belief, defiend as, 

\begin{equation}
\forall a \in A, \ \forall x_a \in E_a, \quad  
b_a(x_a) \propto e^{-H_a(x_a)} 
\prod_{i \in a} \prod_{\substack{b \in A \\ b \neq a, \ i \in b}} m^*_{b \to i}(x_i).
\end{equation}

and 
\begin{equation}
   \forall a \in I, \ \forall x_i \in E_i, \quad  
   b_i^t(x_i) \propto e^{-H_i(x_i)} 
\end{equation}

Then, for any $\alpha \in \mathcal{H}$, $b_\alpha > 0$.  
Fixed points of the belief propagation algorithm are in correspondence with critical points of the generalized Bethe free energy \cite{yedidia2005, PeltrePhD, sergeantperthuis2025functorialitybeliefpropagationalgorithms};  
therefore, $b$ is a critical point of $BT_{\mathcal{A},\mathcal{H}}$.

\end{proof}

\section{Background on the topology of partially ordered sets}

\subsection{Partially ordered spaces as topological spaces}
A poset $\mathcal{A}$ is a set with a binary relation, denoted as $\cdot \leq \cdot$, that is transitive ($b \leq a$ and $c \leq b$ then $c \leq a$), reflexive ($a \leq a$), and antisymmetric ($a \leq b$ and $b \leq a$ then $a = b$). A subposet of $\mathcal{A}$ is a subset $\mathcal{B}$ equipped with the same relation $\leq$. An order preserving map (or isotone map) $\phi: \mathcal{A} \to \mathcal{B}$ between two posets $\mathcal{A}$ and $\mathcal{B}$ is a function such that whenever $b \leq a$ then $\phi(b) \leq \phi(a)$. We will call a function $\phi: \mathcal{A} \to \mathcal{B}$ decreasing when $\phi(a) \leq \phi(b)$ whenever $b \leq a$. In what follows, when referring to maps between two posets we implicitly mean nondecreasing ones. A finite poset can be seen as a finite topological space \cite{stong1966finite}, for the topology generated by neighborhoods $U_a = \{b : b \leq a\}$ (see Appendix \ref{poset-space}); the open subsets $U \subseteq \mathcal{A}$ are the lower-sets, i.e., sets such that whenever $a \in U$ then if $b \leq a$ then $b \in U$; we denote the associated topological space as $X_\A$. In fact, the reverse is true: any finite topological space that is $T_0$ is homeomorphic to a poset \cite{stong1966finite}. Continuous functions from $X_{\mathcal{A}} \to X_{\mathcal{A}}$ are in correspondence with order preserving maps from $\mathcal{A} \to \mathcal{B}$.

There are two standard ways to make a partially ordered set~\(\A\) into a topological space, which we will denote by \(X_\A\) and \(X^\A\).

$X_\A$ is the topological space whose points are the elements of $\A$, and the topology $\mathcal{O}$ on $X_\A$ is generated by the basis of neighborhoods $U_a = \{b \in \A \mid b \leq a\}$ for each $a \in \A$. One says that a subset $U \subseteq \A$ is a \emph{lower set} whenever, for any $a \in U$ and any $b \in \A$ such that $b \leq a$, it follows that $b \in U$. The set of open subsets of $X_\A$ is exactly the collection of lower sets of $\A$.

\subsection{Galois connection}

Loosely speaking, a Galois connection is a map between two posets \(\A\) and \(\B\) that admits a kind of inverse.
 There are several ways to present Galois connections, depending on whether one chooses an order or its opposite, and whether one starts from $\A$ or from $\B$. In the following definition of Galois connection, we adopt a convention that highlights it as a specific case of an adjunction between two categories (the posets seen as categories). In Appendix~\ref{appendix:galois-connection}, we discuss how to relate our convention to the one in~\cite{WALKER1981373}, and why they give rise to equivalent definitions.
\begin{defn}[Galois Connection]\label{galois-connection}
Let $\A, \B$ be two finite posets, and let $g:\A\to \B$ and $f:\B\to \A$ be two order preserving maps such that:

\begin{equation}
\forall a\in \A, \, b\in \B, \quad g(a)\leq b \iff a\leq f(b).
\end{equation}

Then we say that $f,g$ form a Galois connection between $\A$ and $\B$ and denote it as $g \dashv f$. 
\end{defn}

\begin{prop}[Theorem 4.1 \cite{WALKER1981373}]\label{retraction-coeff}
Let $\A, \B$ be two posets, and let $(g:\A\to \B, f:\B\to \A)$ be a Galois connection between $\A$ and $\B$, i.e., $g \dashv f$. Then,  

\begin{equation}\label{theom:eq1}
\forall a \in \B, b \in \A, \quad \sum_{b' : f(b') = b} \mu_\B(a, b') = \sum_{a' : g(a') = a} \mu_\A(a', b).
\end{equation}

\end{prop}

Therefore, 

\begin{equation}\label{theom:eq2}
c_\A(a) = \sum_{a' \,:\, f(a') = a} c_\B(a').
\end{equation}

\begin{proof}
Equation \ref{theom:eq1} is a restatement of Theorem 4.1 of \cite{WALKER1981373}. To prove equation \ref{theom:eq2}, sum on the left and right hand of Equation \ref{theom:eq1} by $\sum_{a\in \B }$, 

$$\sum_{a\in \B}\sum_{b' : f(b') = b} \mu_\B(a, b')=\sum_{a\in \B} \sum_{a' : g(a') = a} \mu_\A(a', b)$$
then, 

$$\sum_{b' : f(b') = b} \sum_{a\in \B}\mu_\B(a, b')= \sum_{a'\in \A} \mu_\A(a', b)= c_\A(b)$$

Therefore, $\sum_{b' : f(b') = b}c_\B(b')= c_\A(b)$.

\end{proof}

\subsection{Homotopic posets and core of a poset}

As we have seen in the previous sections, posets can be viewed as topological spaces; hence one can make precise the claim that they are homotopy equivalent (as topological spaces); this is what we do in the next definition. We also recall the notion of a deformation retract, since we use it extensively in the main result of this paper.

\begin{defn}[Homotopy of maps and posets]
Two order preserving maps $f, g: \mathcal{A} \to \mathcal{B}$ are said to be homotopic if there is a continuous map $h: X_{\mathcal{A}} \times [0,1] \to X_{\mathcal{B}}$ such that for any $a \in \mathcal{A}$, $h(a,0) = f(a)$ and $h(a,1) = g(a)$; we then denote $f\simeq g$; the topology on $[0,1]$ is the usual Euclidean one.  

Let $\mathcal{B}$ be a subposet of $\mathcal{A}$ and denote $i: \mathcal{B} \hookrightarrow \mathcal{A}$ as the inclusion. $\mathcal{B}$ is a deformation retract of $\mathcal{A}$ if there is a map $r: \mathcal{A} \to \mathcal{B}$ such that:

\begin{itemize}
    \item (retraction condition) $r \circ i = \id$
    \item (homotopy condition) $i \circ r \simeq \id$
\end{itemize}

A strong deformation retract imposes that the homotopy $h: X_{\mathcal{A}} \times [0,1] \to X_{\mathcal{A}}$ leaves $\mathcal{B}$ invariant, i.e., for all $b \in \mathcal{B}$ and $t \in [0,1]$, $h(b, t) = b$.

More generally, two posets $\mathcal{A}, \mathcal{B}$ are said to be homotopic if there is a pair of order preserving maps $\phi: \mathcal{A} \to \mathcal{B}$, $\psi: \mathcal{B} \to \mathcal{A}$ such that $\psi \circ \phi \simeq \id$ and $\phi \circ \psi \simeq \id$.

\end{defn}

In \cite{stong1966finite}, a classification of posets up to homotopy is given; it requires introducing \emph{linear} and \emph{colinear} points, also known as up-beat and down-beat points (see \cite{barmak2011algebraic} Section 1.3). The homotopy type of a poset $\A$—that is, whether two posets are homotopic or not—is fully characterized by its core. The core of a poset is a subposet, denoted $co\A$, that is homotopy equivalent to $\A$ and has minimal cardinality. Furthermore, the retractions of linear and colinear points are the two operations that allow one to reduce a poset to its core through a sequence of homotopy equivalences.

\begin{defn}[Linear and colinear points]
    Let $\A$ be a finite poset. Then, $a\in \A$ is linear when there is $a\uparrow> a$, such that  
\begin{equation}\tag{L}\label{def:L}
    \forall b\in \A, \quad b\geq a \implies b\geq a\uparrow.
\end{equation}

    Dually, a point $a$ is colinear when there exists $a\downarrow <a$ such that  

\begin{equation}\tag{coL}\label{def:coL}
    \forall b\in \A, \quad b\leq a \implies b\leq a\downarrow.
\end{equation}
\end{defn}

\begin{lem}\label{lem:homotopy}
Let $\A, \B$ be two finite posets, and let $f, g : \A \to \B$ be two order-preserving maps. If for all $a \in \A$, we have $f(a) \leq g(a)$, then $f$ is homotopic to $g$.
\end{lem}

\begin{proof}
The posets $\A$ and $\B$ can be viewed as finite topological spaces endowed with the lower Alexandrov topology (generated by lower sets), Corollary 3 of\cite{stong1966finite} therefore applies.
\end{proof}

\begin{prop}\label{prop:linear-colinar-adjoint}
Let $\A$ be a finite poset and let $a$ be a linear point. Denote by 
\[
r_{a\uparrow} : \A \to \A \setminus \{a\}
\]
the retraction that sends any $b \neq a$ to itself and sends $a$ to $a\uparrow$. Then \( r_{a\uparrow} \) is a deformation retraction of \( \A \) onto \( \A \setminus \{a\} \).

Similarly, let \( a \) be a colinear point of \( \A \), and let 
\[
r_{a\downarrow} : \A \to \A \setminus \{a\}
\]
be the retraction that sends \( b \neq a \) to itself and sends \( a \) to \( a\downarrow \). Then \( r_{a\downarrow} \) is a deformation retraction of \( \A \) onto \( \A \setminus \{a\} \).

Furthermore, when \( a \) is a linear point, \( r_{a\uparrow} \) is left adjoint to the inclusion map \( i : \A \setminus \{a\} \to \A \), i.e., \( r_{a\uparrow} \dashv i \). When \( a \) is colinear, the inclusion map is right adjoint to the deformation retraction \( r_{a\downarrow} \), i.e., \( i \dashv r_{a\downarrow} \).
\end{prop}

\begin{proof}
In both cases, when \( a \) is a linear or colinear point, we have \( r_{a\uparrow} \circ i = \id \) and \( r_{a\downarrow} \circ i = \id \). By definition, \( i \circ r_{a\uparrow} \geq \id \) and \( i \circ r_{a\downarrow} \leq \id \). Therefore, by Lemma~\ref{lem:homotopy}, \( i \circ r_{a\uparrow} \) is homotopic to \( \id \), and \( i \circ r_{a\downarrow} \) is also homotopic to \( \id \). 

This shows that \( r_{a\uparrow} \) and \( r_{a\downarrow} \) are both deformation retractions onto \( \A \setminus \{a\} \).

The fact that $r_{a\uparrow}\dashv i$ and $i \dashv r_{a\downarrow}$ is a consequence of Proposition \ref{in-appendix:galois-adjoint} Appendix \ref{appendix:galois-connection}. Let us explain the proof, 
following Definition \ref{galois-connection} Appendic \ref{appendix:galois-connection}, when $a$ is a linear points, Equation \ref{def:L} can be restated as,

\begin{equation}
\forall b\in \A\setminus\{a\},\quad  r_{a\uparrow}(a)\leq b\iff a\leq i(b)
\end{equation}

For $b\neq a$, $r_{a\uparrow}(c)=c$ and $c\leq b$ is of course equivalent to $c\leq b$. Therefore $r_{a\uparrow}\dashv i$. 

When $a$ is a colinear point, Equation \ref{def:coL} can be restated as, 

\begin{equation}
\forall b\in \A\setminus\{a\}, \quad i(b)\leq a \iff b\leq r_{a\downarrow}(a)
\end{equation}

Therefore $i\dashv r_{a\downarrow}$.
\end{proof}

\begin{defn}[Core of a poset \cite{stong1966finite}]
The core of a poset $\A$ is a sub-poset $\B\subseteq \A$ such that $\B$ has no linear or colinear points (in $\B$) and such that $\B$ is a strong deformation retract of $\A$. We will denote the core of $\A$ as $co\A$.
\end{defn}

\begin{prop}\label{core-poset-obtained}
Any finite poset $\mathcal{A}$ has a core, and a strong deformation retract $r: \mathcal{A} \to co\A$ is obtained through a sequence of up and down retractions. Two posets are homotopic if and only if their cores are isomorphic.

\end{prop}

\begin{proof}
Proof of Theorem 2 in \cite{stong1966finite}.
\end{proof}

\section{Minima and Critical Points of the Bethe Free Energy under Deformation Retractions}

In this section we explain how one can, by keeping track of the Hamiltonians, reduce optimization of the Bethe free energy over a factor to optimizing it over its core (when regarded as a poset). We do it iteratively by retracting linear and colinear points. 

By a chain in a poset \(\A\), we mean a strictly increasing sequence \(a_1 < a_2 < \cdots < a_n\), with \(a_i \in \A\) for \(i = 1,\dots,n\).

Let $\mathcal{H} = (I,A)$ be a hypergraph on a finite set $I$.  We first remark that any linear point in a factor graph is a minimal element of $\A(H)$: the only chains of $\A(H)$ are constitutes of vertices $i\in I$ and $a\in A$ such that $i<a$ (i.e. $i\in a$) or $i$ itself. The colinear points are maximal elements.  All chains have length at most 1.  After removing a linear or a colinear point, the chains of the resulting subposet are still of length at most 1.

After retracting a linear point \(a\in \A(\mathcal{H})\), it may happen that \(\A \setminus \{a\}\) is no longer the poset associated to any factor graph.  For example, consider the hypergraph $\mathcal{H} = \bigl(I = \{i\},\, A = \{\{i\}\}\bigr)$. Then \(i \subseteq \{i\}\), so \(i\) is a linear point.  After the retraction, the poset contains only the single element \(a = \{i\}\), with \(E_a = E_i\).  But by definition there is no factor graph having \(I = \varnothing\) and \(E_i\) nonempty.  

The previous remark is not a major issue: as explained in Section~\ref{Bethe-free-general}, the Bethe free energy is defined for subcollections \(\A\) of subsets of a finite set \(I\), where for each \(a\in \A\) one defines $E_a = \prod_{i\in a} E_i$. After retraction one still associates to each \(b \in \A \setminus \{a\}\) the finite set $E_b = \prod_{i\in b} E_i$.
The poset obtained after retraction, \(\A \setminus \{a\}\), has maximal chains of length at most 1.

In the next subsection we prove the isomorphism on critical points of the respective Bethe free energies before and after retraction, for finite posets that are collections of subsets of a finite set \(I\) and whose chains have length at most one.

\subsection{Under the retraction of a linear point}

\begin{prop}\label{prop:linear-points-BT}
  Let \(I\) be a finite set, and let \(\A \subseteq \mathcal{P}(I)\) be a poset (under inclusion) whose chains have length at most $1$. Let \(E_i\) for \(i \in I\) be a collection of finite sets and consider a collection of Hamiltonians $\bigl(H_c : E_c \to \mathbb{R}\bigr)_{c \in \mathcal{A}(\mathcal{H})}$. Let $a \in \mathcal{A}$ be a linear point, and set $\mathcal{B} = \mathcal{A} \setminus \{a\}$. Let $E_b=\prod_{i\in b} E_i$, let $H_b : E_b \to \mathbb{R}$ be a collection of Hamiltonians, one for each $b \in \mathcal{A}$.  
Define $i^*H = (H_b)_{b \in \mathcal{B}}$, where \(I\) and \((E_i)_{i \in I}\) remain unchanged.

Then the inclusion $i : \mathcal{A} \setminus \{a\} \hookrightarrow \mathcal{A}$ induces a bijection $\mathbb{L}(\mathcal{A}) \to \mathbb{L}(\mathcal{B})$,  
which restricts to a bijection between the critical points of the Bethe free energy functional $BT_{\mathcal{A},H} : \mathbb{L}(\mathcal{A})^{\circ} \to \mathbb{R}$
and the critical points of the Bethe free energy functional $BT_{\mathcal{B},i^*H} : \mathbb{L}(\mathcal{B})^\circ \to \mathbb{R}$. A similar bijection holds for the global minima over \(\mathbb{L}(\mathcal{A})\) and \(\mathbb{L}(\mathcal{B})\). 
\end{prop}

\begin{proof}

Let us first show that $\mathbb{L}(\mathcal{A})$ is in (linear) bijection with $\mathbb{L}(\mathcal{B})$.  
Define two maps
\[
\phi : \prod_{c \in \mathcal{A}} \mathbb{P}(E_c) \ \to \ \prod_{c \in \mathcal{B}} \mathbb{P}(E_c),
\qquad
s : \prod_{c \in \mathcal{B}} \mathbb{P}(E_c) \ \to \ \prod_{c \in \mathcal{A}} \mathbb{P}(E_c),
\]
defined respectively as follows:

\begin{equation}
\begin{aligned}
&\forall b = (b_c)_{c \in \mathcal{A}}, \ \forall c \in \mathcal{B}, 
&& \phi(b)_c = b_c, \\[6pt]
&\forall b = (b_c)_{c \in \mathcal{B}}, \ \forall c \in \mathcal{B}, 
&& s(b)_c = b_c, 
\qquad \text{and}  \qquad 
\forall x \in E_a, \quad s(b)_a(x) = \sum_{\substack{y \in E_{a\uparrow} \\ y_a = x}} b_{a\uparrow}(y).
\end{aligned}
\end{equation}

Then one shows that : $s \circ \phi = \mathrm{id}$ and $\phi \circ s = \mathrm{id}$.

By Proposition \ref{retraction-coeff} as $a\not\in \im i$, then $c_\A(a)=0$; and for $b\in \A$, such that $b\neq a$, $c_\A(b)= c_\B(b)$. Denote $\left(\mathbb{E}_{Q_b}[H_b] - S(Q_b)\right)$ as $l_b(Q_b)$. \\

Then, for any $Q\in \prod_{b\in \A}\mathbb{P}(E_b)$,

\begin{equation}
\sum_{b\in \A} c_\A(b) l_b(Q_b)=  \sum_{b\in \B}  c_\A(b) l_{b}(Q_{b}).
\end{equation}

And so, for any $Q\in \mathbb{L}(\A)$, 
\begin{equation}\label{equation:isomo1}
 \mathcal{L}_\A(Q)= \mathcal{L}_\B(\phi(Q)) 
\end{equation}

In particular, since $\phi$ is a bijection, the global minima of $\mathcal{L}_\A$ are mapped bijectively to those of $\mathcal{L}_\B$. Let us now derive a similar result for critical points in $\mathbb{L}(\A)^\circ$ and $\mathbb{L}(\B)^\circ$.

We will now differentiate Equation~\ref{equation:isomo1} to obtain the desired property regarding the critical points of the two functionals.

Let us denote $T\mathbb{L}(\A)$ the underlying vector space of the affine space $\mathbb{L}(\A)$ which is the set, 

\begin{equation}
\begin{split}
T\mathbb{L}(\A) = \bigl\{\, u = (u_a)_{a \in \A} \in \prod_{a\in \A} \R^{E_a}
\;\bigm|\; & \forall\, c \le b,\; \forall\, x_c \in E_c,\; u_c(x_c)
= \sum_{\substack{y_b \in E_b \\ y_c = x_c}} u_b(y_b), \\
& \forall a\in\A,\; \sum_{x \in E_a} u_a(x) = 0 \bigr\}.
\end{split}
\end{equation}

For any $u\in T\mathcal{L}(\A)$, and $Q\in \mathcal{L}(\A)^\circ$,

\begin{equation}
    d_Q\mathcal{L}_\A(u)= d_{\phi(Q)}\mathcal{L}_\B\circ d_Q\phi(u)
\end{equation}

\(\phi\) is an isomorphism, so its differential is invertible, and the vanishing of $ d_Q\mathcal{L}_\A$ is equivalent to that of \(d_{\phi(Q)} \mathcal{L}_\mathcal{B}\), which completes the proof.

\end{proof}

\subsection{Under retraction of colinear points}

\begin{defn}\label{psi-qui-def-bij}
  
 Let \(I\) be a finite set, and let \(\A \subseteq \mathcal{P}(I)\) be a poset (under inclusion) whose chains have length at most $1$. Let $a$ be a colinear point of $\A$. Let $(H_c : E_c \to \mathbb{R};c \in \A)$ be a collection of Hamiltonians. Let $\B = \A \setminus \{a\}$ and $i$ be the inclusion $\B\hookrightarrow \A$. Define $\psi^{a,H} : \mathbb{L}(\B) \to \mathbb{L}(\A)$, denoted simply as $\psi$ when clear from context, as the map defined as follows: for any $Q\in \mathbb{L}(\B) $,

\begin{align}
\forall b \neq a \quad & \psi(Q)_b = Q_b, \\
\forall x \in E_a,\quad & \psi(Q)_a(x) = \pi(x \mid x_{a\downarrow})\, Q_{a\downarrow}(x_{a\downarrow}).
\end{align}
where, for any $x_a\in E_a$,
\begin{equation}
\pi\bigl(x_a \mid x_{a\downarrow}\bigr)
= \frac{e^{-H_a(x_a)}}
       {\displaystyle\sum_{\substack{z\in E_a \\ z_{a\downarrow} = x_{a\downarrow}}} e^{-H_a(z)}}.
\end{equation}

\end{defn}

\begin{prop}\label{central-result-colin}
 Let \(I\) be a finite set, and let \(\A \subseteq \mathcal{P}(I)\) be a poset (under inclusion) whose chains have length at most $1$. Let \(E_i\) for \(i \in I\) be a collection of finite sets and consider a collection of Hamiltonians $\bigl(H_c : E_c \to \mathbb{R}\bigr)_{c \in \mathcal{A}(\mathcal{H})}$. Let $a$ be a colinear point of $\A$. Let $\B = \A \setminus \{a\}$ and $i$ be the inclusion $\B\hookrightarrow \A$. Let,

\begin{equation}
\forall Q \in \mathbb{L}(\B), \quad 
\mathcal{L}(Q) = \sum_{b \in \B} c_\B(b) \left(\mathbb{E}_{Q_b}[H_b] - S(Q_b)\right) 
-\mathbb{E}_{Q_{a\downarrow}} \left[ \ln \sum_{\substack{z\in E_a \\ z_{a\downarrow} = X_{a\downarrow}}}
e^{-H_a(z) + H_{a\downarrow}(X_{a\downarrow})}\right]
\end{equation}

Then, $\psi$ induces an isomorphism between the the critical points (respectively, the minima when they exists) of $\mathcal{L} : \mathbb{L}(\B)\to \mathbb{R}$ and the critical points (respectively the minima when they exist) of $BT_{\A, H} : \mathbb{L}(\A) \to \mathbb{R}$.

\end{prop}

\begin{proof}

Let $g:\A\to \B$ and $f:\B\to \A$ such that $g,f$ for a Galois connection, i.e. $f\dashv g$. By Proposition \ref{retraction-coeff},

\begin{equation}
\forall Q \in \prod_{a \in \A} \mathbb{P}(E_a), \quad \sum_{a \in \A} c_\A(a) \left[\mathbb{E}_{Q_a}[H_a]-S(Q_a) \right]  = \sum_{a' \in \B} \sum_{a \in \A\,:\, g(a) = a'} c_\A(a) \left[\mathbb{E}_{Q_a}[H_a]-S(Q_a) \right].
\end{equation}

Let $a\in \A$ be a colinear point, let $\B=\A\setminus\{a\}$. Then $c_\A(b)= c_\B(b)$ for any $b\in \B$ such that $b \neq a\downarrow$, and, 

\begin{equation}
c_\B(a\downarrow)= c_\A(a\downarrow)+ c_\A(a)
\end{equation}

Let $g=r_{a\downarrow}:\A\to \B$ be the retraction of $a$ on $a\downarrow$. The next step is to apply the chain rule for entropy to the joint random variables \((X,Y)\),  
where for \(x \in E_a\) we set \(X(x) = x\) and \(Y(x) = x_{a\downarrow}\). Let $Q_a\in \mathbb{P}(F_a)$, let us denote $Q_{a\downarrow} := \pi^a_{a\downarrow,*} Q_a$ (the marginalisation induced by the projection onto $E_{a\downarrow}$: $\pi^a_{a\downarrow}$) and let,
\[
\forall x_a\in E_a, y_{a\downarrow}\in E_{a\downarrow}, \quad Q_{a \mid a\downarrow}(x_a, y_{a\downarrow}) = \frac{Q_a(x_a)}{Q_{a\downarrow}(y_{a\downarrow})} 1[x_{a\downarrow} = y_{a\downarrow}].
\]  
Then the following holds,
\begin{equation}
\forall Q_a \in \mathbb{P}(E_a), \quad S(Q_a) = S(Q_{a\downarrow})+ \mathbb{E}_{Q_{a\downarrow}} [S(Q_{a \mid a\downarrow})].
\end{equation}

where, more explicitly,
 
\begin{equation}
S(Q_{a \mid a\downarrow})(y_{a\downarrow}) = -\sum_{x_a \,:\, x_{a\downarrow} = y_{a\downarrow}} Q_{a \mid a\downarrow}(x_a, y_{a\downarrow})\ln Q_{a \mid a\downarrow}(x_a, y_{a\downarrow}).
\end{equation}

Therefore, for any $Q_a\in \mathbb{P}(E_a)$,
\begin{align}\label{proof:chainrule}
c_\A(a) &\left(\mathbb{E}_{Q_a}[H_a] - S(Q_a)\right) 
+ c_\A(a\downarrow) \left(\mathbb{E}_{Q_{a\downarrow}}[H_{a\downarrow}] - S(Q_{a\downarrow})\right) \nonumber\\
&=(c_\A(a) + c_\A(a\downarrow)) \left(\mathbb{E}_{Q_{a\downarrow}}[H_{a\downarrow}] - S(Q_{a\downarrow}) \right) + c_\A(a) \left(\mathbb{E}_{Q_a}[H_a - H_{a\downarrow} \circ \pi^{a}_{a\downarrow}] 
- \mathbb{E}_{Q_{a\downarrow}} [S(Q_{a \mid a\downarrow})] \right).
\end{align}
\underline{First part: proof for global minima.}\\

Let us define, for any $P \in \mathbb{L}(\B)$,

\begin{equation}\label{sub-loss}
\mathcal{L}(P) = \min_{Q_a \in \mathbb{P}(E_a) \colon \pi^a_{a\downarrow,*}(Q_a) = P_{a\downarrow}} BT_{\A,H}(P, Q_a).
\end{equation}

Then

\begin{equation}
\min_{P\in \mathbb{L}(\A)} BT_{\A,H}(p)= \min_{P\in  \mathbb{L}(\B)} \mathcal{L}(P)
\end{equation}

One remarks that,

\begin{align}
\forall P \in \mathbb{L}(\B), \quad 
\mathcal{L}(P) =& \sum_{b \in \B} c_\B(b) \left(\mathbb{E}_{P_b}[H_b] - S(P_b)\right)\nonumber\\ 
&+ c_{\A}(a)\left( \min_{\substack{
    Q_a \in \mathbb{P}(E_a) \\
    \pi^a_{a\downarrow,\ast}(Q_a) = P_{a\downarrow}
}}
\mathbb{E}_{Q_a}[H_a - H_{a\downarrow} \circ \pi^{a}_{a\downarrow}] 
- \mathbb{E}_{Q_{a\downarrow}} [S(Q_{a \mid a\downarrow})] \right).
\end{align}

In particular as $a$ is maximal in $\A$ then $c(a)=1$. The sub-optimization problem to solve is the following:
\begin{equation}\label{sub-optim-pb}
\min_{Q_a \in \mathbb{P}(E_a) \colon \pi^a_{a\downarrow,\ast}(Q_a)=P_{a\downarrow}}
\mathbb{E}_{Q_a}[H_a - H_{a\downarrow} \circ \pi^{a}_{a\downarrow}] 
- \mathbb{E}_{P_{a\downarrow}} [S(q_{a \mid a\downarrow})] 
\end{equation}

The parametrization of the constraint is given by, for any $x_a \in E_a$,  with $x_{a\downarrow}= \pi^a_{a\downarrow}(x_a)$,
\[
Q_a(x_a) = \pi(x_a \mid x_{a\downarrow}) P_{a\downarrow}(x_{a\downarrow}).
\]

In particular, for any $x\in E_a$ and $y\in E_{a\downarrow}$, $\pi(x \mid y) =Q_{a \mid a\downarrow}(x_a, x_{a\downarrow}) 1[x_{a\downarrow}=y]$. Pose $f_a(x_a)=H_a(x_a) - H_{a\downarrow}(\pi^{a}_{a\downarrow}(x_a))$, in other words $f_a= H_a-H_{a\downarrow}\circ\pi^{a}_{a\downarrow}$, then, 

\begin{equation}\label{sub-loss-decompose}
\mathbb{E}_{Q_a}[f_a] 
- \mathbb{E}_{P_{a\downarrow}} [S(Q_{a \mid a\downarrow})] = \sum_{y\in E_{a\downarrow}}P_{a\downarrow}(y)\sum_{x_a \,:\, \pi^a_{a\downarrow}(x_a) = y} \pi(x_a\vert y)\ln \frac{\pi(x_a\vert y)}{e^{-f_a(x_a)}}
\end{equation}

In Equation \ref{sub-loss-decompose}, for each $y\in E_{a\downarrow}$, the map 

$$\pi(\cdot\vert y)\to \sum_{x_a \,:\, \pi^a_{a\downarrow}(x_a) = y} \pi(x_a, y)\ln \frac{\pi(x_a\vert y)}{e^{-f_a}(x_a)}$$

is minimal for, 
\begin{equation}
\forall x \in E_a, \quad \pi^*(x \mid y) = 
\frac{e^{-f_a(x)} }
{\displaystyle\sum_{z \in E_a : z_{a\downarrow} = y} e^{-f_a(z)}}\cdot 1[x_{a\downarrow} = y].
\end{equation}

Denote by \( \mathbb{P}_{>0}(E_a \mid E_{a\downarrow}) \) the set of kernels \( y \in E_{a\downarrow} \to \pi(\cdot \mid y) \in \mathbb{P}(E_a) \) such that:

\begin{itemize}
    \item[1)] if \( x_{a\downarrow} \neq y \), then \( \pi(x \mid y) = 0 \).
    \item[2)] for all \( x \in E_a \), \( \pi(x \mid x_{a\downarrow}) > 0 \).
\end{itemize}

$C:\pi\in \mathbb{P}_{>0}(E_a \mid E_{a\downarrow}) \to \mathbb{E}_{Q_a}[f_a] 
- \mathbb{E}_{P_{a\downarrow}} [S(Q_{a \mid a\downarrow})]$ is a strictly convex function bounded by below on $\mathbb{P}_{>0}(E_a \mid E_{a\downarrow})$, therefore it's minimum is unique and attained. Therefore the global minimum of $C$ is attained for $\pi^*:E_{a\downarrow}\to \mathbb{P}(E_a)$ defined as, for any $x\in E_a$ and $y\in E_{a\downarrow}$,

\[
\pi^*(x \mid y) = 
\frac{e^{-H_a(x)}}
{\displaystyle\sum_{z \in E_a : z_{a\downarrow} = y} e^{-H_a(z_a)} }\cdot 1[x_{a\downarrow} = y].
\]

The associated optimal \( Q_a^* \) is such that for any \( x \in E_a \), $Q_a^*(x) = \pi^*(x \mid x_{a\downarrow}) \cdot P_{a\downarrow}(x_{a\downarrow})$. Therefore $(P,Q_a^*)= \phi(Q)$.

In particular, 

\begin{align}
\max_{Q_a \in \mathbb{P}(E_a) \colon \pi^a_{a\downarrow,\ast}(Q_a)= P_{a\downarrow}}
\mathbb{E}_{Q_a}[H_a - H_a \circ \pi^{a}_{a\downarrow}] 
+ \mathbb{E}_{P_{a\downarrow}} [S(Q_{a \mid a\downarrow})] 
&= -\mathbb{E}_{P_{a\downarrow}} \left[ \ln \sum_{z\in E_a \colon z_{a\downarrow} =X_{a\downarrow}} 
e^{-H_a(z) + H_{a\downarrow}(X_{a\downarrow})} \right] \nonumber\\
&=-\sum_{y\in E_{a\downarrow}} P_{a\downarrow}(y)\ln \sum_{z\in E_a \colon z_{a\downarrow}=y} 
e^{-H_a(z) + H_{a\downarrow}(y)} 
\end{align}

We have solved the sub-optimization problem and therefore obtained an analytic expression for $\mathcal{L}$,

\begin{equation}
\forall Q \in \mathbb{L}(\B), \quad 
\mathcal{L}(Q) = \sum_{b \in \B} c_\B(b) \left(\mathbb{E}_{Q_b}[H_b] - S(Q_b)\right) 
-\mathbb{E}_{Q_{a\downarrow}} \left[ \ln \sum_{z \colon z_{a\downarrow} =X_{a\downarrow}} 
e^{-H_a(z) + H_{a\downarrow}(X_{a\downarrow})}\right]
\end{equation}

Therefore, \( \phi \) induces an isomorphism between the minima of \( \mathcal{L} \) and \( BT_{\mathcal{A}, H} \).\\

\underline{Second part: proof for critical points.}\\

Recall that 

\begin{align}
\forall Q \in \mathbb{L}(\A), \quad 
BP_{\A,H}(Q) =& \sum_{b \in \B} c_\B(b) \left(\mathbb{E}_{Q_b}[H_b] - S(Q_b)\right)\nonumber\\ 
&+ c_{\A}(a)\left(\mathbb{E}_{Q_a}[H_a - H_{a\downarrow} \circ \pi^{a}_{a\downarrow}] 
- \mathbb{E}_{Q_{a\downarrow}} [S(Q_{a \mid a\downarrow})]\right) .
\end{align}

The constraint \( Q \in \mathbb{L}(\mathcal{A})^\circ \) can be rewritten as \( (Q_b)_{b \in \mathcal{B}} \in \mathbb{L}(\mathcal{B})^\circ \), and \( \pi^a_{a \downarrow} Q_a = Q_{a \downarrow} \). The second constraint can be reparameterized as
\[
\forall x \in E_a, \quad Q_a(x) = \pi(x \mid x_{a \downarrow}) Q_{a \downarrow}(x_{a \downarrow}).
\]

with \( \pi \in \mathbb{P}_{>0}(E_a \mid E_{a \downarrow}) \). Therefore, \( \mathbb{L}(\mathcal{A})^\circ \cong \mathbb{L}(\mathcal{B})^\circ \times \mathbb{P}_{>0}(E_a \mid E_{a \downarrow}) \). Both \( \mathbb{L}(\mathcal{B})^\circ \) and \( \mathbb{P}_{>0}(E_a \mid E_{a \downarrow}) \) are open subsets of affine spaces, and the underlying vector space of their product, i.e. \( T\mathbb{L}(\mathcal{A}) \), is the product \( T\mathbb{L}(\mathcal{B}) \times T \mathbb{P}_{>0}(E_a \mid E_{a \downarrow}) \). For any $(v,w)\in T\mathbb{L}(\mathcal{A})$, and $Q=(P,Q_a)\in \mathbb{L}(\A)^{\circ}$,

\begin{align}
d_Q BT_{\mathcal{A},H}(v,w) &= d_P BT_{\mathcal{B}, H}(v) \nonumber \\
&\quad + \sum_{y \in E_{a \downarrow}} v_{a \downarrow}(y) \sum_{\substack{x \in E_a \\ x_{a \downarrow} = y}} \pi(x \mid y) \ln \frac{\pi(x \mid y)}{e^{-f_a(x)}} \nonumber \\
&\quad + \sum_{y \in E_{a \downarrow}} P_{a \downarrow}(y) \sum_{\substack{x \in E_a \\ x_{a \downarrow} = y}} w(x \mid y) \left( \ln \frac{\pi(x \mid y)}{e^{-f_a(x)}} + 1 \right) 
\end{align}

Therefore $d_Q BT_{\mathcal{A},H}(v,w)=0$ for all $(v,w)\in T\mathcal{L}(\B)\times T\mathbb{P}_{>0}(E_a\vert E_{a\downarrow})$ is equivalent to, for any $v,w$,

\begin{align}
d_P BT_{\B, H}(v)+ \sum_{y\in E_{a\downarrow}} v_{a\downarrow}(y) \sum_{x \,:\, x_{a\downarrow}= y} \pi(x\vert y)\ln \frac{\pi(x\vert y)}{e^{-f_a(x)}}=0 \label{proof:eq1}\\
 \sum_{x\in E_a \,:\, x_{a\downarrow} = y} w(x\vert y)\left(\ln \frac{\pi(x\vert y)}{e^{-f_a(x)}}+1\right)=0\label{proof:eq2}
\end{align}

The solution to Equation \ref{proof:eq2} is given by, 

\begin{equation}
\forall y\in E_{a\downarrow},\forall x\in E_a,\quad \pi^*(x\vert y)= \frac{e^{-f_a(x)}}{\sum_{x \,:\, x_{a\downarrow}= y}e^{-f_a(x)}}\cdot 1[x_{a\downarrow}=y],
\end{equation}

and Equation \ref{proof:eq1} can be rewritten as, for any $v\in T\lim \mathbb{L}(\B)$,

\begin{equation}
d_P \mathcal{L}(v)=0
\end{equation}

Then $\psi$ is an isomorphism between the critical points of $BT_{\A,H}$ and $\mathcal{L}$.

\end{proof}

In the last part of this section, we rewrite \( \mathcal{L} \) as \( BT_{\mathcal{B}, \widetilde{H}} \) with an appropriate choice of Hamiltonians \( \widetilde{H} \).

\begin{prop}\label{colinear-proof-BT}

Let \(I\) be a finite set, and let \(\A \subseteq \mathcal{P}(I)\) be a poset (under inclusion) whose chains have length at most $1$. Let \(E_i\) for \(i \in I\) be a collection of finite sets and consider a collection of Hamiltonians $\bigl(H_c : E_c \to \mathbb{R}\bigr)_{c \in \mathcal{A}(\mathcal{H})}$. Let $a$ be a colinear point of $\A$. Let \(\mathcal{B} = \mathcal{A} \setminus \{a\}\), and let \(i : \mathcal{B} \hookrightarrow \mathcal{A}\) be the inclusion map. There are two possible cases:
\begin{enumerate}
    \item if $b=a\downarrow$ is a linear point of $\B$, then pose for any $c\in \B$ such that $c\neq b\uparrow$, $\widetilde{H}_b= H_b$ and 
  \begin{align}
\forall x \in E_{b}, \quad 
\widehat{H}_{b}(x) &= \ln \sum_{z\in E_a \colon z_{b} = x} 
e^{-H_a(z) + H_{b}(x)} \nonumber \\
\widetilde{H}_{b\uparrow} &= H_{b\uparrow} - \widehat{H}_{b} \circ \pi^{b\uparrow}_{b}
\end{align}

    \item if $a\downarrow$ is not a linear point of $\B$, then, for any $b\in \B$ such that $b\neq a\downarrow$, $\widetilde{H}_b= H_b$, and 
    
\begin{equation}
\forall x \in E_a,\quad
\widetilde{H}_{a_{\downarrow}}(x)
= H_{a_{\downarrow}}(x)
\;-\; \frac{1}{c_{\mathcal{B}}(a_{\downarrow})}
\left[
  \ln \sum_{\substack{z \in E_a \\ z_{a\downarrow} = x}}
    e^{-H_a(z) + H_{a_{\downarrow}}(x)}
\right].
\end{equation}

\end{enumerate}

Then, $\psi$ induces an isomorphism between the minima of $BT_{\B,\widetilde{H}}: \mathbb{L}(\B) \to \mathbb{R}$ and the minima the of $BT_{\A, H} : \mathbb{L}(\A) \to \mathbb{R}$; and also the critical points of $BT_{\B,\widetilde{H}}: \mathbb{L}(\B)^\circ  \to \mathbb{R}$ and $BT_{\A, H} : \mathbb{L}(\A)^{\circ} \to \mathbb{R}$.
\end{prop}
\begin{proof}

Any element $a \in \A$ is either be a minimal element or a maximal element (or both). In particular, if $a$ is a linear point, then $a\uparrow > a$, so $a$ is a minimal element; conversely, if $a$ is a colinear point, then $a\downarrow < a$, so $a$ is a maximal element.
Let $a\in \A$ be a minimal element but not a linear point, then, there are at least two elements $b,b_1$ such that $a<b$ and $a<b_1$; one shows that, $\mu_{\A}(a,a)=1$ and for any $a<c$ ($c$ is a cover of $a$, i.e. there is not $a<x<c$) $\mu(c,a)=-1$. Therefore $c_\A(a)<-1$. 
Therefore, when $a\downarrow$ is not a linear point of $\B$, by Proposition \ref{central-result-colin}, the critical points and the minima of $BT_{\A,H}$ are in correspondence with those of $\mathcal{L}$, where

\begin{equation}
\forall P \in \mathbb{L}(\B), \quad 
\mathcal{L}(P) = \sum_{b \in \B} c_\B(b) \left(\mathbb{E}_{P_b}[H_b] - S(P_b)\right) 
- \mathbb{E}_{P_{a\downarrow}} \left[ \ln \sum_{z\in E_a \colon z_{a\downarrow} = X_{a\downarrow}} 
e^{-H_a(z) + H_{a\downarrow}(X_{a\downarrow})} \right]
\end{equation}

This expression can be rewritten as
\begin{equation}
\forall Q \in \mathbb{L}(\B), \quad 
\mathcal{L}(Q) = \sum_{b \in \B} c_\B(b) \left(\mathbb{E}_{Q_b}[\widetilde{H}_b] - S(Q_b)\right) = BT_{\B, \widetilde{H}}(Q).
\end{equation}

Assume now that $b=a\downarrow$ is a linear point of $\B$, then by Proposition \ref{retraction-coeff}, $c_\B(b)=0$. $b\uparrow$ is a maximal element of $\B$, therefore $c_\B(b\uparrow)=1$. Therefore,

\begin{align}
\forall Q \in \mathbb{L}(\B), \quad 
\mathcal{L}(Q) =& \sum_{d \in \B:d\neq b\uparrow} c_\B(d) \left(\mathbb{E}_{Q_d}[H_d] - S(Q_d)\right) \nonumber\\
&+\mathbb{E}_{Q_{b\uparrow}}[H_{b\uparrow}] - \mathbb{E}_{Q_{a\downarrow}} \left[ \ln \sum_{z\in E_a\colon z_{a\downarrow} = X_{a\downarrow}} 
e^{-H_a(z) + H_{a\downarrow}(X_{a\downarrow})} \right]- S(Q_a)
\end{align}

Pose for all $x\in E_{a\downarrow}$, 

\begin{equation}
\widehat{H}_{a\downarrow}(x) = \ln \sum_{z\in E_a \colon z_{a\downarrow} = x} 
e^{-H_a(z) + H_{a\downarrow}(x)}
\end{equation}

\

One remarks that, 
\begin{align}
\mathbb{E}_{Q_{b\uparrow}}[H_{b\uparrow}]
- \mathbb{E}_{Q_{a\downarrow}} [\widehat{H}_{a\downarrow}]
&= \mathbb{E}_{Q_{b\uparrow}} \left[ H_{b\uparrow}(X_{b\uparrow}) - \widehat{H}_{a\downarrow} \circ \pi^{b\uparrow}_{a\downarrow}(X_{b\uparrow}) \right] \nonumber \\
&= \sum_{x \in E_{b\uparrow}} Q_{b\uparrow}(x) \left[ H_{b\uparrow}(x) - \widehat{H}_{a\downarrow}(x_{a\downarrow} \right]
\end{align}

Therefore by posing, for $c\in \B$ such that $c\neq  b\uparrow$, $\widetilde{H}_c=H_c$ and $\widetilde{H}_{b\uparrow}= H_{b\uparrow} - \widehat{H}_{a\downarrow} \circ \pi^{b\uparrow}_{a\downarrow}$, then $\mathcal{L}= BT_{\B,\widetilde{H}}$.

\end{proof}

\section{The main result: minima and critical points under deformation retractions of hypergraphs}

\begin{thm}\label{main-result}
Let $\mathcal{F}$ be a factor graph, with sets \(E_i\) for \(i \in I\), hypergraph \(\mathcal{H} = (I, A)\) and Hamiltonians $\bigl(H_c : E_c \to \mathbb{R}\bigr)_{c \in \mathcal{A}(\mathcal{H})}$ (associated to the factors). Denote $\A(\mathcal{H})$ as $\A$. Let $\mathrm{co}\A$ be the core of $\A$. Then, there is a a collection of functions $(\widetilde{H}_c : E_c \to \mathbb{R})_{c \in \mathrm{co}\A}$ that induces an isomorphism between the critical points of $BT_{\A,H}$ and those of $BT_{\mathrm{co}\A,\widetilde{H}}$.
\end{thm}

\begin{proof}
By Proposition \ref{core-poset-obtained}, there exists a sequence of up and down retractions, corresponding respectively to the deletion of linear and colinear points of $\A$, which, when composed, define a deformation retraction from $\A$ onto $\mathrm{co}\A$. Let us denote $a_1,...a_n$ the set of points of $\A$ that are successively deleted to reach $\mathrm{co}\A$. Each $a_k$, for $k=1...n$, are linear or colinear points in their respective posets.

If $a_1$ is a linear point, one can apply Proposition \ref{prop:linear-points-BT}. Denote $\A_1 = \A \setminus \{a_1\}$, and let $i: \A_1 \hookrightarrow \A$ be the inclusion of $\A_1$ into $\A$; let $\phi: \mathbb{L}(\A) \to \mathbb{L}(\B)$ be the induced map. Denote $H_1 = i^* H$; then $\phi$ induces an isomorphism between the critical points of $BT_{\A, H}$ and those of $BT_{\A_1, H_1}$. Denote by $g_{a_1}$ the restriction of $\phi$ to the respective sets of critical points.

If $a_1$ is a colinear point, then one instead applies Proposition \ref{colinear-proof-BT}. In this case, let $H_1 = \widetilde{H}$. Recall that $\psi^{a_1, H}$ (Definition \ref{psi-qui-def-bij}) induces an isomorphism between the critical points of $BT_{\A, H}$ and those of $BT_{\A_1, H_1}$; denote by $g_{a_1}$ the inverse of $\psi^{a_1, H}$, restricted to the respective critical points. The chains of $\A$ are of length at most $1$, so are those of $\A_1$.

Let us build by recursion the sequences $\A_k$, $H_k$, and $g_{a_k}$ for $1 < k \leq n$. Repeating the same argument as for $a_1$, and depending on whether $a_k$ is a linear or colinear point, we set $\A_{k+1} = \A_k \setminus \{a_k\}$; the chains in $\A_{k+1}$ are of length at most $1$ as are those of $\A_k$. $H_{k+1}$ is respectively $i^*H_k$ or $\widetilde{H}_k$; and $g_{a_{k+1}}$ is respectively the restriction of $\phi : \mathbb{L}(\A_k) \to \mathbb{L}(\A_{k+1})$ to the critical points of $BT_{\A_k, H_k}$ and $BT_{\A_{k+1}, H_{k+1}}$, or the restriction of $\psi^{a_k, H_k}$ to those critical points.

Then $g_{a_n}\circ...\circ g_{a_1}$ defines a isomorphism between the critical points of $BT_{\A,H}:\mathbb{L}(\A)\to \R$ and those of $BT_{\mathrm{co}\A,H_{n+1}}: \mathbb{L}(\mathrm{co}\A)\to \R$. 

\end{proof}

\section*{Acknowledgements}

 We are grateful to Ping Xu for valuable discussions.

This research benefited from the support of the FMJH ‘Program Gaspard Monge for optimization
and operations research and their interactions with data science’, and from the support of EDF.

\bibliographystyle{plain}
\bibliography{arxiv}

\appendix

\section{Galois Connection and Adjunction}\label{appendix:galois-connection}

\begin{defn}[A poset as a category]
Consider a finite poset $\A$; the associated category, denoted as $\textbf{A}$
 or simply as $\A$, has as objects the elements $x\in \A$ and a unique morphism $x\to y$ whenever $x\leq y$. 
\end{defn}

\begin{defn}[Galois Connection]
Let $\A, \B$ be two finite posets, and let $g:\A\to \B$ and $f:\B\to \A$ be two  mean nondecreasing maps such that:

\begin{equation}
\forall a\in \A, \, b\in \B, \quad g(a)\leq b \iff a\leq f(b)
\end{equation}

then we say that $f,g$ form a Galois connection between $\A$ and $\B$.
\end{defn}

\begin{rem} Definition \ref{galois-connection} is equivalent to saying that $g$ is left adjoint to $f$ for $\A$ and $\B$ seen as categories. It is the reason why we denoted it in the body of the article $g \dashv f$ when $f,g$ define a Galois connection. The relation to the Galois connection as defined in \cite{WALKER1981373} is that $f,g$ form a Galois connection in our sense if and only if $f,g$ form a Galois connection between $\A^{op}$ and $\B$ according to the conventions of \cite{WALKER1981373}.

\end{rem}

\begin{prop}
Let $\A, \B$ be two finite posets, and let $g:\A\to \B$ and $f:\B\to \A$ form a Galois connection between $\A$ and $\B$, i.e., $g \dashv f$. Then $\A$ and $\B$ have the same homotopy type.
\end{prop}

\begin{proof}
Section 4 \cite{WALKER1981373}
\end{proof}

\begin{prop}\label{in-appendix:galois-adjoint}
A point $a$ is linear if and only if the inclusion map $j: \A\setminus \{a\}$ admits a left adjoint, denoted as $r$. In other words, there exists a unique map $r: \A\setminus \{a\} \to \A$ such that $(r,j)$ is a Galois connection between $\A$ and $\A\setminus \{a\}$. 

Similarly, a point $a$ is colinear if and only if the inclusion map $j: \A\setminus \{a\}$ admits a right adjoint, i.e., if $(j,r)$ is a Galois connection between $\A$ and $\A\setminus \{a\}$.  
\end{prop}

\begin{proof}
If \( a \) is a linear point of \( \mathcal{A} \), then for any \( b \neq a \), we have \( j \circ r(b) = b \) and \( j \circ r(a) = a \uparrow \geq a \geq a \); furthermore, \( r \circ j = \operatorname{id} \). 

Therefore, if for \( b \in \mathcal{A} \) and \( c \in \mathcal{A} \setminus \{a\} \) such that \( r(b) \leq c \), then 
\[
b \leq j \circ r(b) \leq j(c),
\]
and so \( b \leq j(c) \). 

Now assume that \( b \leq j(c) \), then \( r(b) \leq r \circ j(c) = c \); therefore, \( r \dashv j \). Similarly, assume \( a \) is a colinear point of \( \mathcal{A} \), then it is a linear point of \( \mathcal{A}^{op} \). Applying the previous result, it follows that 
\[
r(a) \leq_{op} b \iff a \leq_{op} f(b),
\]
which corresponds to 
\[
b \leq_{op} r(a) \iff f(b) \leq_{op} a,
\]
and so \( j \dashv r \).

Now, assume that \( r \dashv j \), then \( a \uparrow = r(a) \leq b \) if and only if \( a \leq b \), and so \( a \leq b \implies a \uparrow \leq b \); therefore, \( a \) is a linear point. A similar proof shows that \( j \dashv r \) implies that \( a \) is a colinear point. 
\end{proof}

\begin{rem}
Let $a$ be a linear point. Then, the left adjoint to the inclusion $j:\A\setminus\{a\} \to \A$, denoted as $r^{a\uparrow}:\A \to \A\setminus\{a\}$, sends $a$ to $a\uparrow$ and is the identity map on the other points.  

Similarly, let $a$ be a colinear point. Then, the right adjoint to the inclusion $j:\A\setminus\{a\} \to \A$, denoted as $r^{a\downarrow}:\A \to \A\setminus\{a\}$, sends $a$ to $a\downarrow$ and is the identity map on the other points.  
\end{rem}

\section{Partially ordered spaces as topological spaces}\label{poset-space}

There two standard way to make a partially ordered set $\A$ into a topological spaces that we will denote as $X_\A$ and $X^\A$.

$X_\A$ is the topological space whose points are the elements of $\A$, and the topology $\mathcal{O}$ on $X_\A$ is generated by the basis of neighborhoods $U_a = \{b \in \A \mid b \leq a\}$ for each $a \in \A$. One says that a subset $U \subseteq \A$ is a \emph{lower set} whenever, for any $a \in U$ and any $b \in \A$ such that $b \leq a$, it follows that $b \in U$. The set of open subsets of $X_\A$ is exactly the collection of lower sets of $\A$.

$X^\A$ is the space $\A$ equipped with the topology $\mathcal{O}_1$, generated by the neighborhood bases $\{b \in \A \mid a \leq b\}$ for each $a \in \A$. The set of open subsets of $\mathcal{O}_1$ is the set of upper sets, that is, the subsets $U$ such that if $a \in U$ and $b \in \A$ with $a \leq b$, then $b \in U$.

The two spaces $X_\A$ and $X^\A$ are related as follows: $X_\A = X^{\A^{\mathrm{op}}}$.

\end{document}